\documentclass{article}

\usepackage{amsmath}
\usepackage{amssymb}
\usepackage{graphicx}
\usepackage{subcaption}
\usepackage{booktabs}
\usepackage{multirow}
\usepackage{threeparttable}
\usepackage{hyperref}
\usepackage{cleveref}
\usepackage{enumitem}

\newcommand{\ts}{\hspace*{0.4em}}

\usepackage[final]{corl_2022}

\title{Self-Supervised Pre-Training of 3D Point Cloud Networks with Image Data}

\author{
  Andrej Janda,\ts Brandon Wagstaff,\ts Edwin G. Ng,\ts and Jonathan Kelly \\
  Institute for Aerospace Studies\\
  University of Toronto \\
  Canada\\
  \texttt{<first name>.<last name>@robotics.utias.utoronto.ca} \\
}

\begin{document}
\maketitle


\vspace*{-2mm}
\begin{abstract}
    Reducing the quantity of annotations required for supervised training is vital when labels are scarce and costly.
    This reduction is especially important for semantic segmentation tasks involving 3D datasets that are often significantly smaller and more challenging to annotate than their image-based counterparts.
    Self-supervised pre-training on large unlabelled datasets is one way to reduce the amount of manual annotations needed.
    Previous work has focused on pre-training with point cloud data exclusively; this approach often requires two or more registered views.
    In the present work, we combine image and point cloud modalities, by first learning self-supervised image features and then using these features to train a 3D model.
    By incorporating image data, which is often included in many 3D datasets, our pre-training method only requires a single scan of a scene.
    We demonstrate that our pre-training approach, despite using single scans, achieves comparable performance to other multi-scan, point cloud-only methods.
\end{abstract}

\keywords{Self-Supervised Learning, Contrastive Learning, Point Clouds}

\section{Introduction}
\label{sec:introduction}

The two most common representations used for robotic scene understanding tasks are \emph{images} and \emph{point clouds}.
Images are dense and feature-rich, but their lack of depth information limits how well they can be used alone to model 3D environments.
Although point clouds circumvent many of the limitations inherent to images, they are notoriously hard to annotate.
This annotation difficulty is a key limiting factor for many state-of-the-art data-driven scene understanding algorithms that require large, annotated datasets~\citep{he2017mask,charles2017PointNet,jiang2020pointgroup,choy20194d}.
Generating labels requires human annotators to manipulate the clouds by zooming, panning, and rotating to select points of interest.
Annotators then have to separate points that belong to a particular object from the background and other occluded points.

The difficulty of annotating point clouds has resulted in considerable effort and labelling times for existing datasets.
For example, the SemanticKITTI dataset~\citep{behley2019semantic}, which has 518 square tiles of 100 metres length each, required 1,700 hours to label.
ScanNet~\citep{dai2017scannet}, which has 1,600 reconstructed scenes of indoor rooms, took about 600 hours to label~\citep{zhang2021self}.
Despite the substantial labelling time, 3D datasets are still significantly
smaller than comparable image-only datasets.

The labelling effort required for 3D data is the reason we seek, herein, to reduce the volume of annotations necessary.
Previous work~\citep{xie2020pointcontrast, hou2021Exploring, zhang2021self} has demonstrated that \emph{self-supervised contrastive pre-training} is an effective approach for improving performance on scene understanding tasks with raw unlabelled point cloud data.
A key limitation of existing 3D pre-training methods is that they neglect the information-rich images that are often available as part of 3D datasets.
We propose a pre-training method that leverages images as an additional modality, by learning self-supervised image features that can be used to pre-train a 3D model.
Our learning method is split into two stages.
The first stage (Stage 1) learns image features using a self-supervised contrastive learning framework.
The second stage (Stage 2) applies the same contrastive learning framework to pre-train a 3D model by making use of the 2D features learned in Stage 1.
By incorporating visual data into the pre-training pipeline, we obtain a notable advantage: only a single point cloud scan and the corresponding image are required during pre-training.
The use of a single scan obviates the need for two or more overlapping 3D views, which are required by many point-only approaches.
The use of a single scan improves the scalability of our approach, since we require raw 3D data only, as opposed to multiple scans that have been aggregated using a robust mapping pipeline for data association.

Through extensive experimentation, we compare our pre-training approach with existing point cloud-only approaches on several downstream tasks and across several datasets.
We find that our method performs competitively with methods that use multiple overlapping point cloud scans, despite having access to single scans and images only.
In short, we make the following contributions:

\begin{itemize}[leftmargin=0.25in]
    \item we describe a self-supervised method for extracting visual features from images and using them as labels to pre-train 3D models via a contrastive loss;
    \item we provide visualizations demonstrating that the features capture structure, such as lines and surface patches, present in the input image and point cloud;
    \item we demonstrate that a model trained using features learned from raw images improves performance on 3D segmentation and object detection tasks.
\end{itemize}

\section{Related Work}
\label{sec:related-work}

Our approach builds upon work on self-supervised contrastive learning using images and point clouds.
These techniques typically produce two augmented versions of each data point by applying a series of transformations with different sets of parameters.
Subsequently, a contrastive loss minimizes the distance, specified by a given metric, between model outputs of the augmented \emph{positive} pair while maximizing the distance of the pair to other `dissimilar' data points, referred to as \emph{negative} samples.
Most algorithms implement the contrastive loss as an InfoNCE (Info Noise Contrastive Estimation) loss function~\citep{oord2018representation}, as defined by \Cref{eq:contrastive_loss} below.
Here, we provide a brief summary of algorithms that use this approach to pre-train image and point cloud networks.

\paragraph{Contrastive Learning in 2D.}
Self-supervised pre-training using image features has proven to be a successful approach for downstream image-classification tasks, achieving comparable performance to supervised pre-training, as demonstrated by SimCLR~\citep{chen2020simple}.
Variations of this approach, such as using memory banks~\citep{wu2018unsupervied}, momentum encoders~\citep{he2020momentum}, stop-gradients~\citep{grill2020bootstrap,chen2021exploring}, clustering~\citep{caron2020unsupervised,li2021prototypical}, and pseudo-labels~\citep{khosla2021supervised} have also proven effective.
For segmentation tasks, self-supervised pre-training of pixel-level features~\citep{wang2021dense,xie2021propagate,wang2021exploring} offers improved downstream performance when compared to image-level features.

\paragraph{Contrastive Learning in 3D.}
The architectures used for pre-training image networks can be adapted to work with point cloud networks.
PointContrast~\citep{xie2020pointcontrast} is an early example that augments two overlapping 3D scans with random rotations and colour transformations.
Corresponding points between the two transformed scans form a positive pair, while all other points are considered as negative samples in the contrastive loss.
Contrastive Scene Contexts (CSC)~\citep{hou2021Exploring} extends PointContrast with an additional partitioning scheme.
DepthContrast~\citep{zhang2021self} augments a single scan and learns feature vectors at the scan-level instead of at the point-level.

\paragraph{Multimodal Contrastive Learning.}
By projecting 3D points into images, algorithms such as Pri3D~\citep{hou2021pri3d} and SimIPU~\citep{zhenyu2022simipu} leverage 3D data when pre-training models for downstream 2D scene understanding tasks.
In~\citep{hou2021pri3d} and~\citep{zhenyu2022simipu}, the pixel-point pairs that map to the same physical 3D location are used as positive pairs in a (pixel-only) contrastive loss.
Alternatively, pre-training with image data can improve downstream performance on 3D scene understanding tasks.
CrossPoint~\citep{afham2022crosspoint} applies a contrastive learning objective to global scene features generated from synthetic point clouds of computer-modelled objects and corresponding rendered images, while Superpixel-driven Lidar Representations (SLidR)~\citep{sautier2022image} uses point-pixel matches from outdoor driving datasets.

\section{Methodology}
\label{sec:method}

\begin{figure}[t]
    \centering
    \includegraphics[width=13cm]{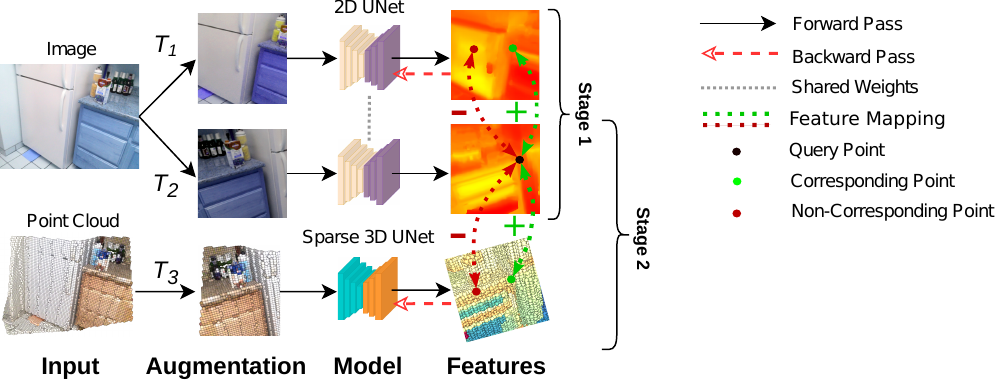}
    \vspace{1mm}
    \caption{Overview of our multimodal contrastive learning framework.}
    \label{fig:overview}
    \vspace{-2mm}
\end{figure}

Our method can be split into two distinct and sequential stages, as shown in \Cref{fig:overview}.
The first stage applies a 2D CNN to generate image features at the pixel level, based on a contrastive loss on the individual pixels.
The second stage then uses these image features to train a 3D model.

\paragraph{Stage 1.}
We utilize the ResUNet architecture from Godard at al.~\citep{godard2019Digging} to extract 2D pixel-level features and modify the decoder to compute a 16-dimensional feature vector for each pixel in the input image.
We pre-load the weights of the encoder from a model trained on the large ImageNet corpus \cite{deng2009imagenet}.
To pre-train the full model, images are selected from a desired pre-training dataset.
We follow roughly the same data augmentation strategy and use the same InfoNCE loss function as SimCLR~\citep{chen2020simple}, except that we compare pixel-level features instead of image-level features.
Pixels that map back to the same coordinates in the original image are considered as positive samples, while all others, including those from other images in a batch, are considered as negative samples.

\paragraph{Stage 2.}
In this stage, we pre-train a 3D point-level feature extraction model using pixel-level features from Stage 1.
We use the 3D model from \cite{xie2020pointcontrast} and treat the final $1\times1$ convolution as the decoder, which we initialize from scratch for training on downstream tasks.
Each point cloud is also augmented so that the model learns to be invariant to differences in orientation, point density and colour fluctuations.
The 2D network is held frozen and the fixed 2D features act as a target for the 3D model to learn.
Mapping between 3D points and 2D features is done via perspective projection.
Each pixel-point match forms a positive pair $(\mathbf{z}_i,\mathbf{z}^+_i)$, where $\mathbf{z}_i$ and $\mathbf{z}^+_i$ represent the feature vectors of a 3D point and the corresponding pixel, respectively.
The feature vector of any other point is considered a negative sample and represented as $\mathbf{z}^-_j$.
We use an InfoNCE loss function, defined as
\begin{equation}
    \mathcal{L} = -\sum_{i = 1}^{N}\log \frac{\exp(\mathbf{z}_{i} \cdot \mathbf{z}^{+}_{i} / \tau)}{\exp(\mathbf{z}_{i} \cdot \mathbf{z}^{+}_{i} / \tau) + \sum^{K}_{j}\exp(\mathbf{z}_{i} \cdot \mathbf{z}^{-}_{j} / \tau)},
    \label{eq:contrastive_loss}
\end{equation}
where $\tau$ is a smoothing parameter and $K$ represents the number of negative samples.
The final parameters of the 3D model are then used on downstream 3D scene understanding tasks.

\section{Experimental Results}
\label{sec:results}

We compare our method to three point-only pre-training methods:
PointContrast~\citep{xie2020pointcontrast}, Contrastive Scene Contexts (CSC)~\citep{hou2021Exploring} and DepthContrast~\citep{zhang2021self}.
We investigate the performance gain of all pre-training algorithms on three downstream tasks: semantic segmentation, instance segmentation, and object detection.
These tasks are applied on three datasets of indoor office environments: ScanNet~\citep{dai2017scannet}, S3DIS~\citep{armeni20163D}, SunRGBD~\citep{song2015sunrgbd, janoch2011category, xiao2013database, silberman2012indoor}; as well as one outdoor driving dataset, \linebreak SemanticKITTI~\citep{behley2019semantic, geiger2012are}.
Pre-training is performed with the ScanNet dataset and all downstream tasks are run using the pipeline and parameters from CSC \cite{hou2021Exploring}.

\begin{figure*}
    \begin{subfigure}[t]{0.27\textwidth}
        \includegraphics[width=3.6cm]{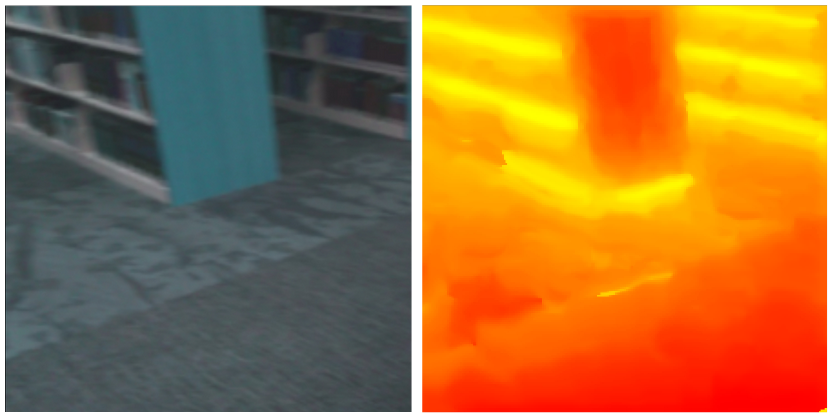}
        \includegraphics[width=3.6cm]{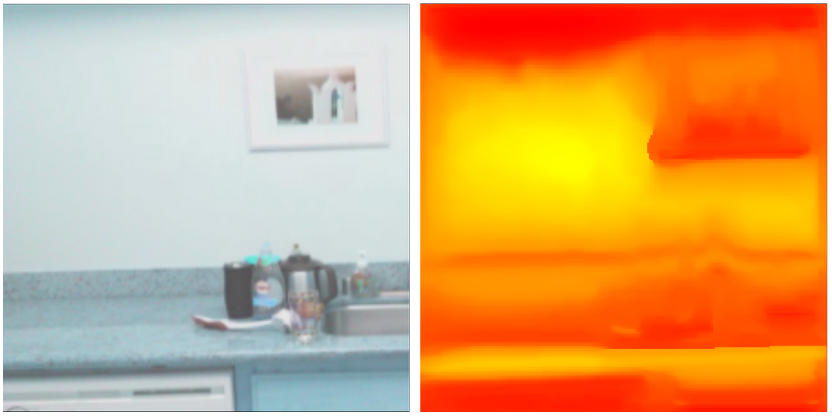}
        \caption{\raggedright
            Input image on the left with corresponding feature visualization on the right.
        }
        \label{fig:features2dvis}
    \end{subfigure}
    \hfill
    \begin{subfigure}[t]{0.27\textwidth}
        \includegraphics[width=3.8cm]{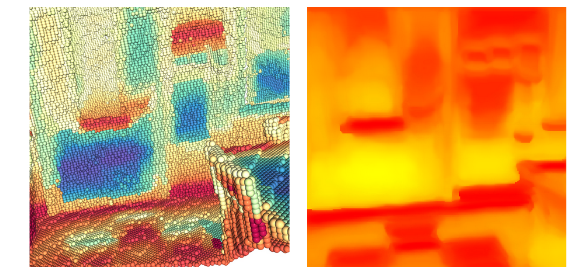}
        \includegraphics[width=3.8cm]{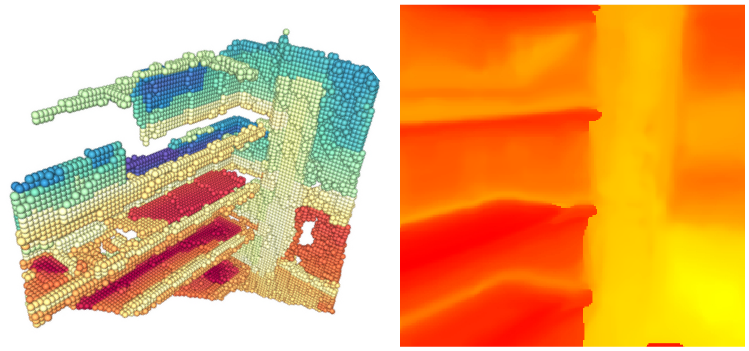}
        \caption{\raggedright
            Point features on the left with corresponding image features on the right.
        }
        \label{fig:features2d-3dvis}
    \end{subfigure}
    \hspace{0.5cm}
    \begin{subfigure}[t]{0.35\textwidth}
        \includegraphics[width=4.9cm]{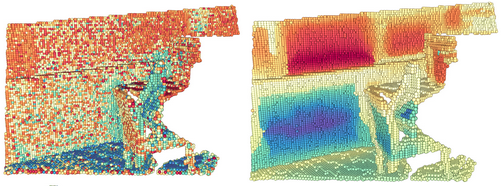}
        \includegraphics[width=4.9cm]{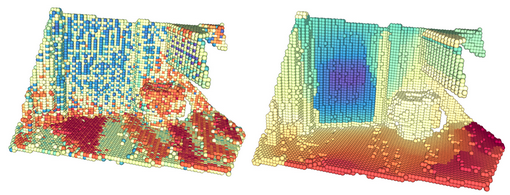}
        \caption{\raggedright
            Point features before (left) and after (right) pre-training using our method.
        }
        \label{fig:FeaturesScratchVsPretrained}
    \end{subfigure}
    \caption{Feature vector visualizations.
    }
    \label{fig:three_graphs}
\end{figure*}

\paragraph{Feature Visualization.}
We verify visually that the 2D features learned using our 2D pre-training scheme have some connection to the original image by following the approach from \cite{choy2019Fully}.
\Cref{fig:features2dvis} shows a comparison between heatmap visualizations and original images, where there is a clear mapping between the input image and the output features.
The heatmaps tend to `highlight' structures such as lines and surface patches that are present in the input images.
\Cref{fig:features2d-3dvis} shows the relation between the 2D and 3D features.
There is a clear mapping between the image and the corresponding point cloud heatmaps.
This correlation verifies that the 3D model has indeed learned to `mimic' the features of the 2D model.
\Cref{fig:FeaturesScratchVsPretrained} displays the difference between randomly initialized and pre-trained features.
Features that are pre-trained follow visible object boundaries.

\paragraph{Downstream Performance.}
A comparison of the downstream performance with different pre-training methods is shown in \Cref{tab:downstreamResults}.
We find that our algorithm is the best performing of the single-scan methods and comparable to multi-scan methods, often outperforming PointContrast.
Interestingly, pre-trained weights on the indoor ScanNet dataset were able to improve performance on the outdoor SemanticKITTI dataset.
However, those same weights offered no improvement when also fine-tuned on ScanNet segmentation tasks, perhaps due to the lack of new information available between pre-training and fine-tuning.
%

\begin{table}[t]
    \centering
    \resizebox{\textwidth}{!}{
        \begin{threeparttable}
            \begin{tabular}{ c c | c c | c c c | c | c}
                \toprule
                \multicolumn{2}{c|}{\multirow{2}{*}{\textbf{Pre-Training Method}}} & \multicolumn{2}{c|}{\textbf{S3DIS}} & \multicolumn{3}{c|}{\textbf{ScanNet}} & \textbf{KITTI}       & \textbf{SUNRGBD}                                                                                                 \\
                                                                                   &                                     & Semantic                              & Instance             & Semantic             & Instance             & Object               & Semantic             & Object               \\
                \midrule
                                                                                   & Scratch                             & 65.1                                  & 53.0                 & 67.4                 & 49.0                 & 35.2                 & 41.0                 & 32.0                 \\
                \midrule
                \multirow{2}{*}{\textbf{Multi-Scan}}                               & PointContrast                       & 66.2 (+1.1)                           & 54.8 (+1.8)          & 66.9 (-0.5)          & 49.1 (+0.1)          & \textbf{36.7 (+1.5)} & 42.1 (+1.1)          & 34.2 (+2.2)          \\
                                                                                   & CSC                                 & \textbf{69.0 (+3.9)}                  & \textbf{57.8 (+4.8)} & \textbf{67.6 (+0.2)} & \textbf{49.3 (+0.3)} & 36.1 (+0.9)          & \textbf{43.0 (+2.0)} & \textbf{35.1 (+3.1)} \\
                \midrule
                \multirow{2}{*}{\textbf{Single-Scan}}                              & DepthContrast                       & 64.9 (-0.2)                           & 52.3 (-0.7)          & 67.4 (+0.0)          & \textbf{48.7 (-0.3)} & 33.9 (-1.3)          & 42.0 (+1.0)          & 32.9 (+0.9)          \\

                                                                                   & \textbf{Ours}                       & \textbf{66.5 (+1.4)}                  & \textbf{55.8 (+2.8)} & \textbf{67.7 (+0.3)} & 48.5 (-0.5)          & \textbf{37.7 (+2.5)} & 42.0 (+1.0)          & \textbf{33.1 (+1.1)} \\
                \bottomrule
            \end{tabular}
        \end{threeparttable}
    }
    \vspace{3mm}
    \caption{Downstream performance comparison of pre-training methods. Semantic segmentation uses the mIOU metric, while both instance segmentation and object detection tasks use the mAP@0.5 metric with a minimum correct overlap ratio of 0.5. The best performance on each task and dataset for the single- and multi-view categories are highlighted in bold.}
    \label{tab:downstreamResults}
    \vspace{-5mm}
\end{table}

\section{Conclusion}
\label{sec:conclusion}

In this short paper, we presented a framework for learning dense pixel-level features from raw unlabelled images, which are then used as targets in a contrastive loss to pre-train a 3D model.
Our visualization showed that the 2D features were successfully 'mimicked' by the 3D model.
We found that our method was the best-performing single-scan method and performed comparably to other multi-scan methods.
These results confirm that incorporating visual data into the pre-training process is a viable strategy to reduce the need for registered point clouds as part of pre-training.
By relaxing this requirement, our method scales more readily when compared to other pre-training techniques that require multiple (registered) scans for contrastive learning.


\clearpage

\bibliography{venues_full,2022-janda-contrastive-corl-prl}

\end{document}